\documentclass[10pt,twocolumn,letterpaper]{article}
\usepackage{wacv}
\usepackage{times}
\usepackage{epsfig}
\usepackage{graphicx}
\usepackage{amsmath}
\usepackage{amssymb}
\usepackage{algorithm}
\usepackage[noend]{algorithmic}
\usepackage{booktabs,subcaption,amsfonts,dcolumn}
\usepackage{float}
\usepackage{multirow}

\usepackage[pagebackref=true,breaklinks=true,letterpaper=true,colorlinks,bookmarks=false]{hyperref}

\wacvfinalcopy % *** Uncomment this line for the final submission

\def\wacvPaperID{319} % *** Enter the wacv Paper ID here

\ifwacvfinal\fi
\begin{document}

%%%%%%%%% TITLE
\title{DVPT: Dynamic Visual Prompt Tuning of Large Pre-trained Models for  Medical Image Analysis}

\author{Along He\textsuperscript{1}, Kai Wang\textsuperscript{1}, Zhihong Wang\textsuperscript{1}, Tao Li\textsuperscript{1}, Huazhu Fu\textsuperscript{2}
\vspace{0.2cm}
\\
\textsuperscript{1}Nankai University\quad\textsuperscript{2}Institute of High Performance Computing (IHPC), Agency for Science, \\Technology and Research (A*STAR), Singapore}

\maketitle
%\thispagestyle{empty}
%%%%%%%%% ABSTRACT
\begin{abstract}
Limited labeled data makes it hard to train models from scratch in medical domain, and an important paradigm is pre-training and then fine-tuning. Large pre-trained models contain rich representations, which can be adapted to downstream medical tasks. However, existing methods either tune all the parameters or the task-specific layers of the pre-trained models, ignoring the input variations of medical images, and thus they are not efficient or effective. In this work, we aim to study parameter-efficient fine-tuning (PEFT) for medical image analysis, and propose a dynamic visual prompt tuning method, named DVPT. It can extract knowledge beneficial to downstream tasks from large models with a few trainable parameters. Firstly, the frozen features are transformed by an  lightweight bottleneck layer to learn the domain-specific distribution of downstream medical tasks, and then a few learnable visual prompts are used as dynamic queries and then conduct cross-attention with the transformed features, attempting to acquire sample-specific knowledge that are suitable for each sample. Finally, the features are projected to original feature dimension and aggregated with the frozen features. This DVPT module can be shared between different Transformer layers, further reducing the trainable parameters. To validate DVPT, we conduct extensive experiments with different pre-trained  models on medical classification and segmentation tasks. We find such PEFT method can not only efficiently adapt the pre-trained models to the medical domain, but also brings data efficiency with partial labeled data.  For example, with 0.5\% extra trainable parameters, our method  not only outperforms state-of-the-art PEFT methods, even surpasses the full fine-tuning by more than 2.20\% Kappa score on medical classification task. It can saves up to 60\% labeled data and 99\% storage cost of ViT-B/16. 
\end{abstract}

\begin{figure}[!t]
	\centering
	\includegraphics[height=2.0in]{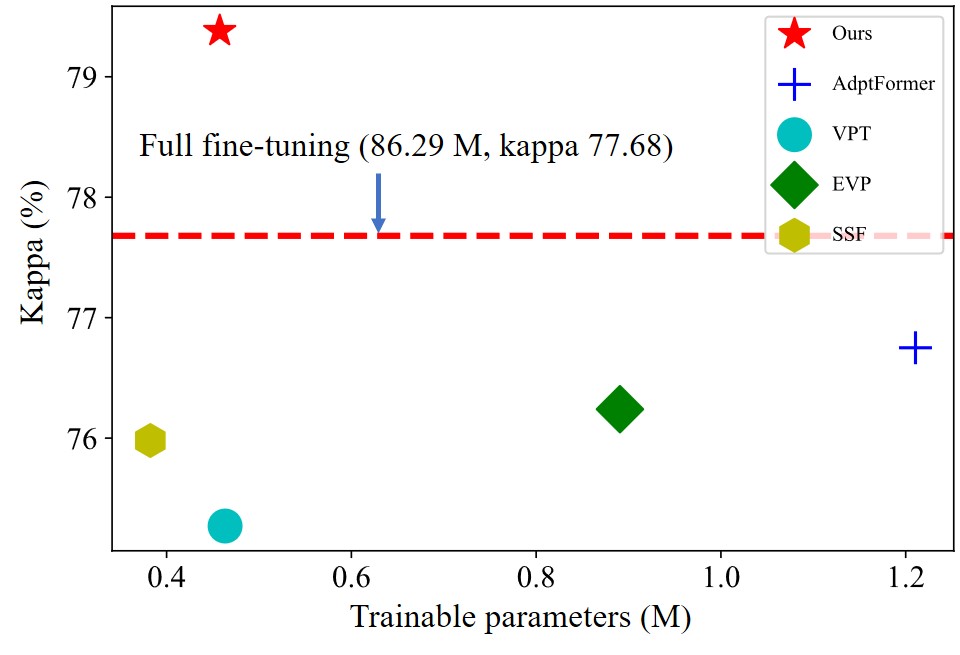}
	%\subfigure{\includegraphics[height=2.5in]{fig/1.pdf}}
	\caption{Performance and trainable parameters trade-off comparison of our method and existing  parameter-efficient tuning methods  \cite{jia2022visual,lianscaling,liu2023explicit,chen2022adaptformer} on the DR grading dataset \cite{li2019diagnostic}. The pre-trained model is ViT-B/16 \cite{dosovitskiy2020image} that  pre-trained on 400 million image-text pairs \cite{radford2021learning}. Our method  achieves better trade-off between performance and parameter efficiency  (higher is better). }
	\label{fig_1}
\end{figure} 

\section{Introduction}
\label{introduction}
Medical image analysis has been the fundamental yet challenging task in computer aided diagnosis, and it has been significantly developed with the rise of deep learning models \cite{ronneberger2015u,zhou2019unet++,chen2021transunet,cao2021swin}. However, the labeled data in the medical domain is limited, which greatly increases the risk of over-fitting when the model is trained from scratch. “Pre-training and then fine-tuning” paradigm is an effective practice.  Recently, large models pre-trained on large-scale datasets have shown amazing capabilities in natural language processing (NLP) \cite{kenton2019bert,ouyang2022training,zhao2023survey} and computer vision \cite{wang2022image,radford2021learning,shen2023hugginggpt}. These pre-trained large models are powerful by fine-tuning them on downstream tasks \cite{hu2021lora,pfeiffer2020adapterfusion} because of  their excellent transfer ability. 

Most of current methods focus on full fine-tuning manner \cite{li2019canet,he2020cabnet,liao2022learning,gu2019net},  where all the parameters of the model are updated. However, fine-tuning the over-parameterized large models can easily lead to over-fitting and  catastrophic forgetting  \cite{Kirkpatrick2016OvercomingCF}, and then cause a significant performance drop, especially for  medical domain that high-quality labeled medical data are often scarce.  Moreover, full fine-tuning of large models for each task  causes  significant tuning computational cost and storage overhead, because each downstream task needs to update and save the entire large model. It is expensive and even infeasible, especially for billion-scale models \cite{dosovitskiyimage,dehghani2023scaling,zhai2022scaling} (\textit{e.g}., ViT-G/14 \cite{zhai2022scaling} with over 1,800 million parameters \textit{vs}. ResNet-50 \cite{he2016deep} with 25 million parameters).

A naive solution for the above problem is fine-tuning only the task-specific layers and then freezing  the other pre-trained parameters. However, this practice lacks  adaptation capacity to downstream tasks and usually yields inferior performance compared to the full fine-tuning. Motivated by the success of the parameter-efficient fine-tuning (PEFT) strategy in the field of NLP \cite{houlsby2019parameter,hu2021lora}, similar approaches have been proposed in the field of computer vision   \cite{he2022parameter,jia2022visual,gao2022visual,lianscaling,liu2023explicit,zhang2020side}. They aim to achieve comparable performance to full fine-tuning with a small number of trainable parameters. Among them, visual prompt tuning (VPT) \cite{jia2022visual} is a promising approach, which uses learnable prompts as inputs and they interact with the image tokens during fine-tuning. These learnable prompts based methods achieved  significant performance improvement compared to fine-tuning only the  task-specific layers, even better than full fine-tuning manner. 

Despite popularity and success of existing works for PEFT \cite{jia2022visual,lianscaling,liu2023explicit,chen2022adaptformer}, they still need large-scale labeled data to fine-tune downstream tasks, which is not feasible in the medical domain with limited labeled data. Compared with natural images, region of interest in medical images usually occupies a small part of the whole image and varies greatly in texture and scale. Therefore, it is necessary for the model to generate sample-specific features according to the intrinsic characteristics of each sample to learn better decision boundaries. While the learned features of existing  PEFT methods  \cite{jia2022visual,lianscaling,liu2023explicit,chen2022adaptformer} are static rather than dynamic, that is, after fine-tuning, their parameters are fixed and each sample is treated without distinction, and they cannot adapt to the complex texture changes of medical images. These methods are seldomly explored in medical domain with limited labeled data, and when these methods are transferred to the field of medical image analysis, the results are not satisfactory and there is still plenty of room for improvement, as shown in Fig. \ref{fig_1}. 

Therefore, how to achieve good performance while fine-tuning a small number of parameters for large pre-trained models is a non-trivial problem in medical domain and it is still an open question. Firstly, we should consider the complex texture of medical images and the scale variation of lesion regions. Secondly, there is a large gap between the pre-trained source domain and target medical domains. Thirdly, there are limited labeled data, and we need to overcome the potential risk of over-fitting and catastrophic forgetting. The key challenges of medical PEFT is how to effectively extract sample-specific and domain-specific knowledge from large  pre-trained models with very few trainable parameters and limited labeled data. Meanwhile, the domain-general knowledge should also be preserved and adapt the pre-trained models to various medical tasks.

In this paper, we aim to study the PEFT method for vision transformers across different medical tasks, and we only fine-tune a few parameters and the remaining parameters are shared.  We tackle the practical yet challenging  problem with the proposed DVPT block, see in Fig. \ref{fig_structure}. Firstly, the original features are transformed through a non-linearly  lightweight bottleneck layer to learn the domain-specific distribution of medical tasks. Then, only a few learnable parameters are regarded as dynamic queries, which are used to do the cross-attention visual prompt tuning (CAVPT) with the transformed features, attempting to acquire sample-specific features in medical domain that are beneficial to medical tasks. Finally, the sample-specific features are further projected to original input dimensions by an expansion layer and combined with original pre-trained domain-general features. The two projection layer with CAVPT are called DVPT, and they can be shared among the pre-trained layers. We only fine tune the shared DVPT block and the task-specific layers, while freezing the pre-trained parameters. It can explore medical features from pre-trained large models in a  parameter-efficient way, which can save computational cost and storage overhead in real-world deployment scenarios. Therefore, the pre-trained models can be shared for different tasks with a small number of task-specific parameters. We can freeze the shared model and efficiently switch different medical tasks by replacing the few task-specific parameters, reducing the storage requirement significantly.

To validate the proposed method, we finetune the pretrained models on medical image classification and segmentation tasks, including both 2D and 3D medical data.  The results show that DVPT can effectively adapt pre-trained models to parameter-efficient medical visual learners with only 0.5\% tunable parameters, achieving  higher performance than previous state-of-the-art (SOTA)  PEFT methods in medical image analysis tasks. Compared to ViT-B/16 with full fine-tuning, our method can reduce the number of trainable parameters by 99\%  and the labeled data requirement by 60\%. It is surprising that parameter efficiency brings data efficiency, and compared with full fine-tuning with 100\% labeled data, DVPT achieves competitive results in the low data regime with only 40\% of labeled data. \textbf{Note that our goal is not to achieve the SOTA performance on specific medical image analysis task, but instead to explore a new paradigm for PEFT in medical domain.}  We hope our findings can inspire and facilitate future research on this topic. Compared with previous approaches, we make the following main contributions:

\begin{itemize}
\item We proposed a simple yet effective framework, namely DVPT, which learns the sample-specific and domain-specific features, reducing the domain gap and adapting large pre-trained vision models to various medical image analysis tasks. The learnable parameters are very few and thus reduce the storage cost significantly.

\item Moreover, our method brings data efficiency. Compared with full fine-tuning with 100\% labeled data, DVPT can achieve comparable results in the low data regime with only 40\% of labeled data for 2D medical data, which not only reduces the number of trainable parameters, but also alleviates the dependence of large models on labeled data.

\item Extensive experiments on both 2D and 3D downstream medical  image classification and segmentation tasks demonstrate the superiority of our framework, which outperforms the current SOTA  PEFT methods with a smaller number of tunable parameters.
\end{itemize}

\section{Related Work}
\label{related}

\subsection{Parameter-efficient Fine-tuning.} 
Parameter-efficient fine-tuning (PEFT) is to inject a few learable parameters into the pre-trained models, and the injected parameters are used to adapt the pre-trained distribution to downstream tasks. The parameters of pre-trained models are frozen to generate general representations learned from large-scale datasets.  PEFT originates from NLP \cite{houlsby2019parameter,hu2021lora,li2021prefix,Zhang2023LLaMAAdapterEF}, and they adopt  text prompt to adapt large language models to downstream tasks. Besides fixed prompt templates, learnable prompts  can be optimized by interacting with the pre-trained models, achieving comparable results. Inspired by NLP, prompt tuning is also explored in computer vision \cite{chen2022adaptformer,liu2023explicit,jia2022visual} and vision-language tasks \cite{Zhou2022ConditionalPL}. VPT \cite{jia2022visual} appends learnable prompts to the pre-trained models, and then optimizes the prompts for the downstream tasks while freezes the pre-trained models. EVP \cite{liu2023explicit} is proposed for low-level structure segmentation, and the tunable parameters focus on the explicit visual content from the features of frozen patch embedding and the input’s high-frequency components.  DePT \cite{gao2022visual} adopts a simple data-efficient prompt tuning method for test-time domain adaptation, which bootstraps the source representation to the target domain by memory bank-based online pseudo-labeling. In order to reduce the inference cost, SSF \cite{lianscaling}  only needs to scale and shift the deep features with a smaller number of tunable parameters, and these parameters can be merged into the original pre-trained model weights via re-parameterization in the inference phase.

In this paper, we propose DVPT and show its promising performance with only a few learnable prompts and it can be shared among pre-trained blocks to learn the sample-specific and domain-specific features for a variety of medical image analysis tasks. Our work shows that current PEFT methods do not fully exploit the capacity of  pre-trained large model in medical domain, as shown in  Fig. \ref{fig_1}. 

\subsection{Pre-trained Large Models}
Early network architecture design in computer vision mainly focused on convolution neural networks, (\textit{e.g.}, ResNet, DenseNet and VGGNet) \cite{he2016deep,huang2017densely,simonyan2014very}. With the tremendous success of Transformers \cite{vaswani2017attention} in NLP \cite{kenton2019bert,ouyang2022training,wu2023visual}, recently, the Transformer based methods have shown promising performance for vision tasks, like ViT, Swin Transformer \cite{dosovitskiyimage,liu2021swin} and SAM \cite{kirillov2023segment}, which facilitates Transformer in various tasks and shows the trend of big convergence. As the training data scale (\textit{e.g.}, ImageNet-21K and JFT-300M) increase, the scale of vision models have increased significantly, from million-scale to billion-scale \cite{xie2021segformer,kirillov2023segment,zhai2022scaling,dehghani2023scaling,radford2021learning,he2022masked}.  Moreover, with rich semantic information from  web-scale multi-modal image-text pairs, vision-language pre-trained models  \cite{radford2021learning,wang2022image,li2022blip,yaofilip} achieved enormous performance improvements compared to the uni-modal models. These well pre-trained models are highly beneficial for downstream tasks. In today's era of large models, ``pre-training and then fine-tuning'' is gradually becoming a trend for downstream tasks. 

Therefore, how to adapt these pre-trained large models to medical domain with a few learnable parameters is a topic worthy of investigation, especially for scenarios where labeled data is scarce in the medical domain.

\section{Methodology}
\label{method}

	\begin{figure*}
	\centerline{\includegraphics[width=5.8in]{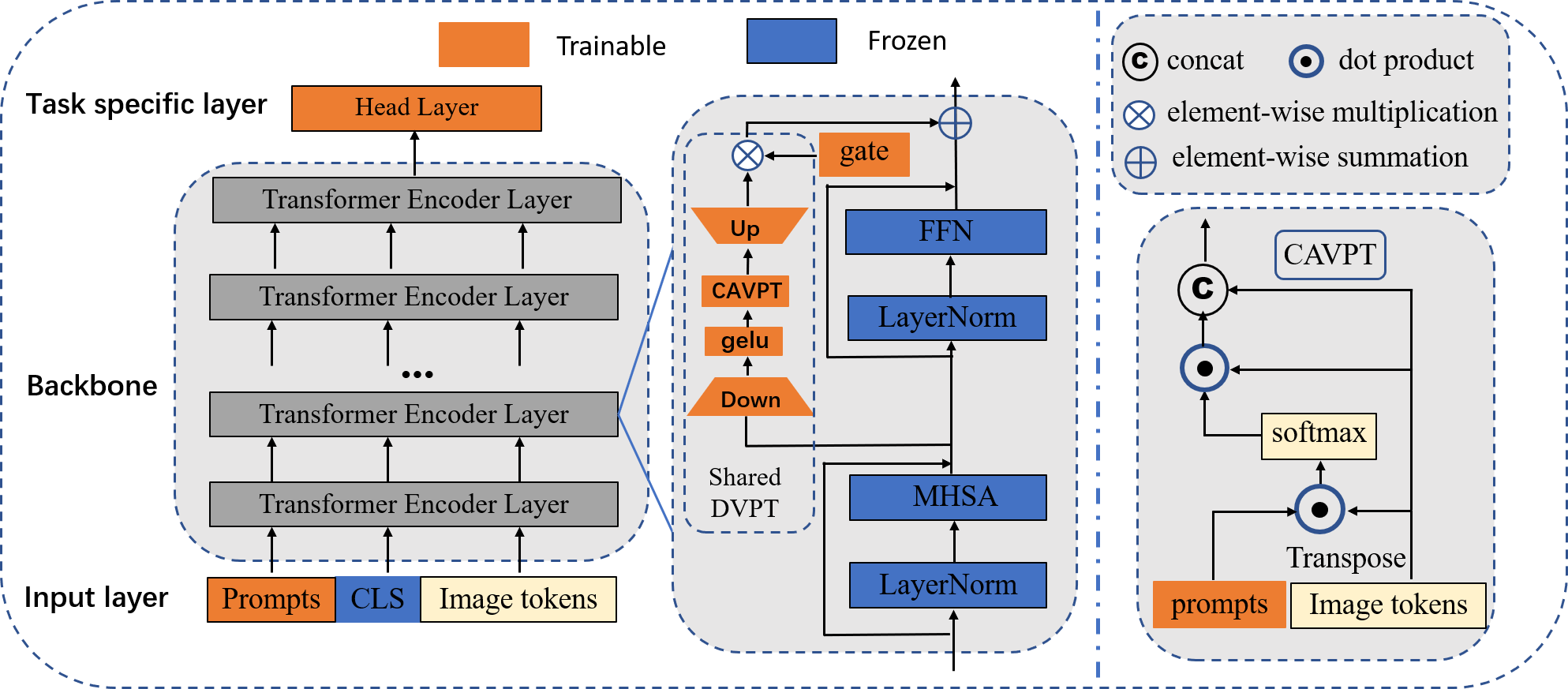}}
	\caption{Overall architecture of the proposed DVPT for parameter-efficient fine-tuning in medical image analysis,  which is base on pre-trained large Vision Transformer. Our proposed method contains two key components: cross-attention visual prompt tuning (CAVPT) and the lightweight projection (down and up projection) layers, and they are trainable together with task-specific layer in the fine-tuning process.}
	\label{fig_structure}
\end{figure*}

In this section, we first present the preliminaries of vision Transformers and overview of our method. And then we introduce the proposed DVPT in detail. The outline of our proposed structure is illustrated in Fig. \ref{fig_structure}. 

\subsection{Preliminaries of Vision Transformers}

We first revisit the widely-used Vision Transformer (ViT) \cite{vaswani2017attention}. Specifically, given an input image $I \in \mathbb{R}^{H\times W\times C}$, $H$, $W$ and $C$ denote the height, width and the channel number of the image, respectively. First, the input image is divided into $n$ patches $Z = \{Z_i \in \mathbb{R}^{p \times p\times C}, i = 1,2,...,n\}$ (the patch size $p=4$ for Swin-transformer \cite{liu2021swin} and $p=16$ for ViT), and $ n = \frac{H\times W}{p^2}$, $Z_i$ is the $i$-th patch. Then, the patches are  projected to image tokens $Z \in \mathbb{R}^{n\times d}$ with a patch embedding  layer.  A class token  $Z_{cls} \in \mathbb{R}^{1\times d}$ is concatenated to the image tokens to form the original token sequence $E = [Z_{cls}, Z]\in \mathbb{R}^{(n+1)\times d}$. Then, they are added with a learnable positional embedding $B \in \mathbb{R}^{(n+1)\times d}$, that is $E_0 = E +B$.  Then, the input tokens $E_0$ are fed into the Transformer blocks. 

For the given input tokens $E_0 = [Z_{cls}, Z]$, they are encoded by $L$ Transformer blocks, and the Transformer block is  defined as follows: 
\begin{equation}
	E'_{l} = \textrm{MHSA}(\textrm{LN}(E_{l-1})) + E_{l-1}, l=1,2,3,...,L,
\end{equation}
\begin{equation}
	E_{l} = \textrm{FFN}(\textrm{LN}(E'_{l})) + E'_{l},l=1,2,3,...,L,
\end{equation}
where $L$ is the number of Transformer encoder blocks, LN is layer normalization and $E_{l}$ is the output of $l$-th Transformer block. MHSA is the multi-head self-attention block, which is formulated as:
\begin{equation}
	\textrm{MHSA}(x) = \psi (Q(x)K(x)^T/\sqrt{d})V(x),
\end{equation}
where $Q\in \mathbb{R}^{d\times d}, K\in \mathbb{R}^{d\times d}, V\in \mathbb{R}^{d\times d}$ are the query, key and value transform matrices, $d$ is the dimension of query and key, and  $\psi$ is the Softmax function to normalize the attention scores to $[0,1]$. FFN is the feed forward network, which is formulated as:
\begin{equation}
	\textrm{FFN}(E'_{l}) =\sigma (E'_{l}W_1) W_2 ,
\end{equation}
where $E'_{l} \in \mathbb{R}^{(n+1)\times d}$ is the features from the self-attention module, $W_2\in \mathbb{R}^{4d\times d}$ and $W_1\in \mathbb{R}^{d\times 4d}$ are the weights of linear layers and $\sigma$ is the GELU \cite{kenton2019bert} nonlinear activation function, respectively. Finally, the class token  $Z_{cls}$ is used for classification tasks and $E_{L}$ is used for segmentation tasks through the head layers:
\begin{equation}
	\hat{y} = \textrm{head}_{cls}(Z_{cls}) \quad or  \quad  \hat{y} = \textrm{head}_{seg}(E_{L}),
\end{equation}

\subsection{Overview of Structure}

We aim to study the PEFT in medical domain with limited data to reduce the tuning computational cost and storage overhead. The key challenge is to improve the learning ability with a few trainable parameters and frozen large pre-trained models. We tackle this problem by utilizing the proposed dynamic visual prompt tuning with cross-attention visual prompt tuning and lightweight projection layers, which can be shared among different layers and provide an effective way to perform feature representation learning with only a few trainable parameters. The overall pipeline of our proposed method is illustrated  in Fig. \ref{fig_structure}. The pre-trained large model is used as the backbone network and their parameters are frozen. We extract useful features for downstream medical tasks from the pre-trained models with the proposed method.

Specifically, we use $F$ to denote the pre-trained model with parameters $\theta$.  For PEFT methods, a few trainable parameters $\theta'$ are inserted into $F$, and the output can be obtained by:
\begin{equation}
	\hat{y} = F(I;\theta,\theta'),
\end{equation}
where $I$ is the input image, and the number of $\theta'$ is usually much less than $\theta$, $\theta$ are frozen during fine tuning, and only $\theta'$ are learnable. As shown in Fig. \ref{fig_structure}, DVPT is paralleled to FFN at each Transformer block, and can be shared among different Transformer blocks. The extracted features are transformed through a non-linearly lightweight layer to learn the domain-specific distribution of medical tasks, and then it adopts learnable prompts as dynamic queries to interact with the domain-specific features by cross-attention, attempting to distill sample-specific representations that are beneficial to downstream tasks. This is more flexible than static prompts after fine-tuning. Finally, to make full use of the domain-general features from the pre-trained model, we projected the sample-specific features to original input dimensions by an expansion layer, and they are added together to complement each other. 

\textbf{Fine-tuning.} During the fine-tuning stage, we only train the extra added parameters and the output head layers to optimize the model and keep other pre-trained parameters fixed.  For  classification tasks, the learnable prompts and class token at the last layer are average-pooled to obtain the final representation, and cross-entropy is employed as the loss function.  For segmentation tasks, only the last output layer of decoder and the extra added parameters  are optimized. Hybrid dice and cross-entropy loss is used for the segmentation loss function. During fine-tuning, we only need to save the  task-specific parameters, which can greatly save large storage space, especially for some edge devices.

\textbf{Inference.} After fine-tuning, we keep the shared parameters, and only load the task-specific trained weights. Therefore, the pre-trained models are able to be adapted to various medical tasks with very few trainable parameters.  The total model size grows slowly when more downstream tasks are added.

\subsection{Dynamic Visual Prompt Tuning}

Different from natural images, the scale and texture of lesions in medical images vary greatly, and thus they have diverse sample-specific features. For previous methods based on trainable prompts, once the prompts are fine-tuned, their weights are fixed at inference time. Therefore, they cannot effectively adapt to the features of downstream samples. We need to design more efficient prompt tuning mechanisms that can generate sample-specific features for different samples. To further learn the domain-specific distribution of downstream medical tasks, we perform the transform for the obtained features and adapt the pre-trained distribution to medical tasks. Inspired by previous works \cite{chen2022adaptformer, liu2023explicit, jia2022visual}, we propose Dynamic Visual Prompt Tuning (DVPT), which includes two lightweight bottleneck layers and the cross-attention visual prompt tuning. The design of DVPT that parallels to the original FFN branch is simple yet effective, which is illustrated in Fig. \ref{fig_structure}.

Specifically, at the input layer of the ViT, we introduce trainable prompt tokens $P \in \mathbb{R}^{m\times d}$,  which are $m$ parameterized vectors of dimension $d$. They are concatenated with the input tokens, and form $E_0^p = [P_0, Z_{cls}, Z_0] \in \mathbb{R}^{(n+m+1)\times d}$, and then, the trainable prompts are propagated through the Transformer encoders. The learnable prompts $P$ can interact with every image and class tokens at each block. 
After interacting with the original image tokens,  a compression layer with weights $f_{down} \in \mathbb{R}^{d\times d'}$ is used for feature transforming the pre-trained features to the medical-specific distribution:
\begin{equation}
	\tilde{E}_{l} = \textrm{MHSA}(\textrm{LN}(E^p_{l-1})) + E^p_{l-1}, l=1,2,3,...,L,
\end{equation}
\begin{equation}
	\hat{E}^d_{l} = \sigma(f_{down}(\tilde{E}_{l})),l=1,2,3,...,L,
\end{equation}
where $\tilde{E}_{l} \in \mathbb{R}^{(n+m+1)\times d}$ are the output features of MHSA and $\hat{E}^d_{l} \in \mathbb{R}^{(n+m+1)\times d'}$ are the bottleneck features. 
%		They can be described as follows:
%	    \begin{equation}
	%			\tilde{E}^p_{l} = \textrm{MHSA}(\textrm{LN}(E^p_{l-1})) + E^p_{l-1}, l=1,2,3,...,L,
	%		\end{equation}
%where $\tilde{E}^p_{l}$ is the fused features of MHSA after adding learnable prompts, which can learn medical information from the original pre-trained large models.
%However, the prompts learned in VPT is static after fine-tuning. They cannot capture sample-specific features and cannot explicitly build the relationship between the prompts and the original image features, and thus they lack feature diversity, which is also the main drawback of VPT.

In order to make the learnable prompts learn sample-specific  features as much as possible,  we propose cross-attention visual prompt tuning (CAVPT) for feature learning. CAVPT is appended to the compression layer to learn sample-specific features, as shown in  Fig. \ref{fig_structure}. CAVPT treats trainable prompts as \textbf{dynamic queries} and uses them to compute similarities with the image features, and the image features are then weighted by the similarities, they are formulated as:
\begin{equation}
	P_{l}, [Z_{cls}, Z_l] = \textrm{split}(\hat{E}^d_{l}), l=1,2,3,...,L,
\end{equation}
\begin{equation}
	P'_{l} = \psi (P_{l}[Z_{cls}, Z_l]^T/\sqrt{d'})[Z_{cls}, Z_l],
\end{equation}
where $split$ denote the splitting of the sequence of length $n+m+1$ into the $n+1$ image and class tokens $[Z_{cls}, Z_l] \in \mathbb{R}^{(n+1)\times d'}$ and $m$ prompt tokens $P_{l}  \in \mathbb{R}^{m\times d'}$.  $\psi$ is the softmax function.
$P'_{l} \in \mathbb{R}^{m\times d'}$ is the sample-specific features, which aggregates the context by softmax attention that relies on prompt-to-token similarities. To preserve the domain-general features and avoid catastrophic forgetting, we concatenate the sample-specific features and domain-general features  for next layers.
\begin{equation}
	\hat{E}^p_{l} =[P'_{l},Z_{cls}, Z_l] , l=1,2,3,...,L,
\end{equation}
Therefore, CAVPT can not only learn sample-specific features, but also retain the rich knowledge of the general domain that share the same features with medical domain. By altering the number of prompts $P$ and the latent dimension $d'$, we can easily control the number of learnable parameters  with different adaptation capacity for downstream tasks.

Finally, an extension layer with weights $f_{up} \in \mathbb{R}^{d'\times d}$ is used to project the features to original dimension:
\begin{equation}
	\hat{E}'_{l} = g \cdot f_{up}(\hat{E}^p_{l}),l=1,2,3,...,L,
\end{equation}
where $g \in \mathbb{R}$ is a learnable gate to control the flow of information dynamically, $\hat{E}'_{l}$ is the output features of the DVPT, which can learn domain-specific and sample-specific features of medical downstream tasks.  Then, it will be fused with the original pre-trained features by residual connection, without forgetting previous domain-general knowledge.:
\begin{equation}
	E_{l} = \textrm{FFN}(\textrm{LN}(\tilde{E}_{l})) + \tilde{E}_{l} + \hat{E}'_{l},l=1,2,3,...,L,
\end{equation}
where $E_{l}\in \mathbb{R}^{(m+n+1)\times d} $ is the output of $l-$th Transformer block after dynamic visual prompt tuning.

Note that DVPT can be shared between every $s$ layers and this mechanism further reduces the number of trainable parameters because the feature distribution is similar between adjacent pre-trained layers, and further reduces the risk of over-fitting. We can decide how many layers to be shared based on the complexity of the downstream tasks.

\subsection{Learnable Parameters Analysis. } 
Our proposed shared DVPT block is parameter-efficient, and the learnable parameters come from DVPT and task-specific head layer. Note that we only add visual prompts at the input layer, and the number of parameters for CAVPT is $m \times d$, for the two projection layers, the number of parameters is $2 \times d \times d'+d+d'$, the hidden dimension $d'$ is a small value compared with $d$, and $d' \ll d$ ($d'=20$, $m=50,d = 768$ in our setting).   Therefore, the total number of parameters are $md'+\frac{L}{s}(2dd'+d+d')$. The parameters of the pre-trained models are frozen, and the number of learnable parameters is small, the number of parameters are 457,446 (0.46M, not shared, $s$ = 1) and 268,414 (0.27M, shared, $s$ = 2) when using ViT-B/16  as pre-trained model. The learnable parameters are about 0.54\% ($s$ = 1) and 0.31\% ($s$ = 2) of the pre-trained model, which is very efficient and effective.

\section{Experiments}
\label{experiments}
In this section, we evaluate the effectiveness of DVPT in both 2D and 3D medical image analysis tasks. We first introduce the experimental setup, and then we compare our method with SOTA PEFT methods and conduct thorough  ablation studies to justify the influence of the two proposed key components. Considering the reproducibility, we report mean results over 3 independent training for all settings and models.

	\subsection{Experimental Settings}

\subsubsection{Dataset}
\textbf{DDR dataset} \cite{li2019diagnostic}. This is a dataset for diabetic retinopathy (DR) grading, which contains 6,835 training images, 2,733 validation images and 4,105 test images. These images are graded into six classes: no DR, mild DR, moderate DR, severe DR, proliferative DR and ungradable. In experiments, we only focus on the five-class  DR grade task following the actual clinical application, and thus ungradable is not used. As a result, the training, validation and test images are 6,320, 2,503 and 3,759 images, respectively. \textbf{Kvasir-SEG dataset} \cite{jha2020kvasir}. This is a dataset for gastrointestinal polyp segmentation, which contains 1000 polyp images and their corresponding binary label masks. Since there is no official train and test dataset split, we perform 5-fold cross validation and report their average results. \textbf{Skin lesion dataset \cite{gutman2016skin}.} This is a dermoscopic image analysis benchmark challenge for automated diagnosis of  skin cancer. The dermoscopy image dataset we used is the ISIC 2016 skin lesion segmentation challenge dataset. It contains 900 training dermoscopic images and 379 test images, and the binary ground truth masks are labeled by the experts. 	\textbf{ACDC dataset} \cite{bernard2018deep}. Automatic Cardiac Diagnosis Challenge (ACDC) is a dataset for automated cardiac diagnosis, which conains 100 3D Magnetic Resonance Imaging (MRI) cases. For all these data, the corresponding manual annnotations for the cavity of the right ventricle (RV), the myocardium (MYO) of the left ventricle and the cavity of the left ventricle (LV) are given by clinical experts. The dataset is split into 70 training samples, 10 validation samples and 20 test samples.  
\textbf{Synapse dataset} \cite{Menze2015TheMB}. This is a dataset for multi-organ segmentation, which consists of 30 CT scans (20 training samples and 10 test samples). There are 13 organs to be segmented, and they are spleen, right kidney, left kidney, gallbladder, esophagus, liver, stomach, aorta, inferior vena cava, portal vein and splenic vein, pancreas, right adrenal gland and left adrenal gland. 
\subsubsection{Pre-trained models.} In the experiments, the pre-trained large models include ViT-B/16 \cite{dosovitskiy2020image} and Swin-base\cite{liu2021swin} for classification task, and SegFormer-B4 \cite{xie2021segformer} for segmentation task. ViT-B/16 is  pre-trained on 400 million image-text pairs with contrastive learning \cite{radford2021learning}, and Swin-base is pre-trained on ImageNet-21K \cite{Deng2009ImageNetAL}.  SegFormer-B4 is pre-trained on ADE20K \cite{Zhou2016SemanticUO}. 

\subsubsection{Implementation Details} 
For the trainable parameters, they are initialized randomly. Horizontal flips, vertical flips, and random crops are applied as data augmentation.  For other compared methods, we follow their default settings from their original papers. The pre-trained Transformer models are trained with the Adam optimizer \cite{kingma2014adam}. For fine-tuning on DDR, the input size is set to 288 $\times$ 288 with batch size of 70, the initial learning rate is 0.01  and  is trained for 80 epochs. For skin lesion and poly segmentation, the input size is set to 512 $\times$ 512 with batch size of 16, and their initial learning rates are 0.003 and 0.01, respectively. For 3D medical data, all the models are trained with 512 $\times$ 512  2D slices and tested in a slice-by-slice fashion for all the 3D volumes, and then the predicted 2D slices are stacked together to reconstruct the 3D prediction results for evaluation.
Note that all the experiments were conducted with the same experiment settings for fair comparisons and our framework is implemented with Pytorch and trained with NVIDIA GeForce RTX 3090 GPUs.  Code will be available at \url{https://github.com/NKUhealong/DVPT}.

\subsubsection{Evaluation Metrics} For DR grading task, the class distribution is imbalanced, following previous work \cite{he2020cabnet}, we adopt the quadratically weighted kappa score \cite{cohen1968weighted} and Accuracy (Acc) to evaluate the DR grading performance.  For segmentation tasks, Dice Similariy Coefficient (Dice),  Intersection-over-Union (IoU) and 95\% Hausdorff distance (HD95) are used for performance evaluation.

\subsection{Comparisons with SOTA Methods}
To demonstrate the effectiveness of our method, we compare it with recent SOTA PEFT methods on both classification and  segmentation medical tasks. To make fair comparisons, we obtain the results under the same settings, and the details of baselines are listed below.

\textbf{Baselines.} We first compare our method with two naive fine-tuning methods: 
(1) full fine-tuning trains all parameters of the models;
(2) linear probe trains only the parameters of the task-specific head  layers. 
We also compare our method with recent PEFT methods: 
(3) VPT \cite{jia2022visual},  the trainable prompts and image tokens are fed into transformers to learn the features. 
(4) SSF \cite{lianscaling}, they introduce the trainable scale and shift factors to modulate the features from the previous operation, and features are performed dot product with a scale factor and then summed with a shift factor.
(5) EVP \cite{liu2023explicit} learns explicit prompts from image embedding and high-frequency components, and perform adapter for each layer.
(6) AdaptFormer \cite{chen2022adaptformer} replaces the original MLP block with AdaptMLP, which consists of two branches, including the frozen branch and the trainable bottleneck module.

\textbf{The results of DR grading task:} We first evaluate the performance of our proposed method on DR grading task against other competitive methods and report the quantitative results in Table \ref{table_sota_cls}.  We can see although linear probe has much fewer trainable parameters than full fine-tuning, its performance is lower than full fine-tuning more than 10\% in Acc and Kappa scores, and the result is not satisfactory. Then, we can find that the performances of recent PEFT methods outperform linear probe, which demonstrates that the transfer ability of large models can be greatly improved by fine-tuning a few trainable parameters.  However, all their performances are worse than those of full fine-tuning, indicating that there is room for further improvement in the medical domain. Finally, our method adopts trainable visual prompts as dynamic queries to learn sample-specific features, and uses projection layer to adapt the pre-trained distribution to downstream medical tasks, and not only outperforms the previous SOTA methods even with less tunable parameters than SOTA PEFT methods, but also beats full fine-tuning method (the theoretical upper bound). We can see that our method still achieves competitive results using the shared DVPT (s=3) with fewer parameters, and thus we can easily control the representation power of the model based on the complexity of the downstream tasks to achieve a good balance between performance and computational cost.

\begin{table} [ht!] \scriptsize
	\center
	\caption{Comparison results of DR grading results with ViT-B/16 model pre-trained on 400 million image-text pairs and swin-base pre-trained on ImageNet-21K. } 
	\setlength{\tabcolsep}{3pt}
	\renewcommand\arraystretch{1.2}
	%\subtable[Comparison results of DR grading results with ViT-B/16 model pre-trained on 400 million image-text pairs. ]{
		\begin{tabular}{l|c|c|cc}
			\hline
			\multirow{2}{*}{Method} & \multirow{2}{*}{Architecture} & \multirow{2}{*}{\#Params(M)}  & \multicolumn{2}{c}{ DDR }\\
			&&&Acc(\%)&Kappa(\%) \\
			\hline
			full fine-tuning  & ViT-B/16      				& 86.2935   & 75.56 & 77.68 \\
			linear probe  & ViT-B/16     				    & 0.0025   & 64.17 & 61.62  \\
			full fine-tuning  & Swin-B      				& 86.7934   & 74.67 & 75.99  \\
			linear probe  & Swin-B      					& 0.0051   & 63.74 & 57.09  \\
			\hline	
			VPT \cite{jia2022visual}       & ViT-B/16      &  0.4633  & 72.60 & 75.27 \\
			SSF \cite{lianscaling}         & ViT-B/16      &  0.3824  & 74.90 & 75.98\\
			EVP \cite{liu2023explicit}     & ViT-B/16      &  0.8907  & 73.34 & 76.24 \\
			AdaptFormer \cite{chen2022adaptformer} & ViT-B/16      &  1.2106  & 74.78 & 76.75 \\
			DVPT (s =3)                   & ViT-B/16      &  0.2053  & 75.52 & 78.83 \\
			DVPT (s =2)                   & ViT-B/16      &  0.2684  & 76.08 & 79.18 \\
			DVPT (s =1)                   & ViT-B/16      &  0.4574  & \textbf{76.40} & \textbf{79.91}\\
			\hline	
			VPT \cite{jia2022visual}       & Swin-B           &  0.3254  & 62.97 & 55.64 \\
			SSF \cite{lianscaling}         & Swin-B       &  0.3790  & 62.06 & 66.45 \\
			EVP \cite{liu2023explicit}     & Swin-B       &  1.5362  & 71.72 & 74.49 \\
			AdaptFormer \cite{chen2022adaptformer} & Swin-B       &  1.5828  & 70.44 & 73.81\\
			DVPT (s =3)                   & Swin-B      &  0.1001 & 74.28 & 77.97 \\
			DVPT (s =2)                   & Swin-B      &  0.1327  & 76.78 & 78.67 \\
			DVPT (s =1)                   & Swin-B       &  0.2196  & \textbf{76.86} & \textbf{80.58} \\
			\hline
		\end{tabular}
		\label{table_sota_cls}
		
		\label{table_sota}
	\end{table}
	
	\begin{table*} [ht!] \scriptsize
		\center
		\caption{Comparison results on lesion segmentation tasks with SegFormer-B4 \cite{xie2021segformer} models pre-trained on ADE20K \cite{Zhou2016SemanticUO}.} 
		\setlength{\tabcolsep}{3pt}
		\renewcommand\arraystretch{1.2}
		\begin{tabular}{l|ccc|cc|cc|cc|cc}
			\hline
			\multirow{2}{*}{Method}  & \multirow{2}{*}{Params(M)} &\multirow{2}{*}{FLOPs(G)}&\multirow{2}{*}{Memory(G)}& \multicolumn{2}{c|}{Polyp  \cite{jha2020kvasir}  } & \multicolumn{2}{c|}{Skin  \cite{gutman2016skin}} & \multicolumn{2}{c|}{Synapse \cite{Menze2015TheMB}}& \multicolumn{2}{c}{ACDC \cite{bernard2018deep}}\\
			&              &  &&        Dice(\%)$\uparrow$ & IoU(\%)$\uparrow$& Dice(\%)$\uparrow$ & IoU(\%)$\uparrow$&Dice(\%)$\uparrow$ & HD95$\downarrow$ &Dice(\%)$\uparrow$ & HD95$\downarrow$\\  
			\hline
			full fine-tuning                                & 61.6383 &55.77& 12.03   & \textbf{91.79}  &\textbf{86.86} &\textbf{92.88}  & \textbf{87.31} & \textbf{74.12}  & \textbf{17.37} & \textbf{88.21}  & \textbf{2.66}\\
			linear probe                  					& 0.0061 &55.77& 4.59    & 61.86  & 49.17 & 84.33  & 75.29  & 39.05  & 38.70 & 54.63  & 23.21\\
			\hline	
			\multicolumn{6}{l}{Parameter-efficient fine-tuning methods}\\
			VPT \cite{jia2022visual}      					&  0.5757 &56.51&10.30   & 55.80  & 42.87 & 83.17  & 74.23 & 26.60  & 41.74 & 54.62  & 19.28\\
			SSF \cite{lianscaling}       					&  0.3125 &55.77&13.85  & 90.13   & 84.67 & 91.26 & 85.48  & 62.43  & 28.65 & 72.36  & 9.35\\
			EVP \cite{liu2023explicit}    					&  0.5566 &57.15& 10.79 & 90.39   & 84.71 & 91.41 & 85.67  & 61.74  & 33.38 & 86.01  & 3.89\\
			AdaptFormer \cite{chen2022adaptformer} 			&  1.5011 &57.99& 10.54  & 91.50  & 85.30 & 91.33  & 85.35  & 63.42  & 25.62 & 80.12  & 4.23\\
			DVPT (s =2)                   	                &  0.1817 &56.34& 9.61 & 91.01   & 86.13 & 92.16 & 86.31  & 64.42  & 23.42 & 86.24  & 4.37\\
			DVPT (s =1)                  					&  0.2940 &56.65& 10.80  & \textbf{91.92}  & \textbf{86.74} & \textbf{92.85}  & \textbf{87.12} & \textbf{65.83}  & \textbf{19.29} & \textbf{87.56}  & \textbf{3.01}\\
			\hline
			\multicolumn{6}{l}{SOTA segmentation methods}\\
			UNet\cite{ronneberger2015u} 					& 23.63 &33.39& 12.50& 87.96   & 82.34 & 89.56   & 83.61  & 67.89   & 26.60 & 89.41   & 4.59\\
			UNet++\cite{zhou2019unet++}                     & 24.38 &35.60& 14.50& 87.01   & 78.84 & 89.46   & 83.63  & \textbf{68.50}   & 42.39 & 89.58   & 4.31\\
			DeepLabv3+ \cite{chen2018encoder}               & 26.19 &33.89& 7.79& 87.04   & 82.05  & 90.12   & 84.35 & 66.53   & 29.58 & 88.25   & 4.21 \\  
			TransUNet\cite{chen2021transunet}               & 109.54 &56.66&7.50 & \textbf{89.21}   & \textbf{83.73} & \textbf{91.31}   & \textbf{84.96}  & 68.04   & 27.21 & 89.71   & \textbf{3.96}\\
			Swin-UNet\cite{cao2021swin}						& 27.27  &36.98&11.92 & 70.71   & 60.96  & 90.12   & 83.21 & 67.74   & \textbf{20.28} & \textbf{90.00}   & 5.30 \\ 					
			\hline
		\end{tabular}
		\label{table_sota_seg}
	\end{table*}
	
	\begin{figure*}[h!]
		\centering
		\subfloat{\includegraphics[width=0.84in]{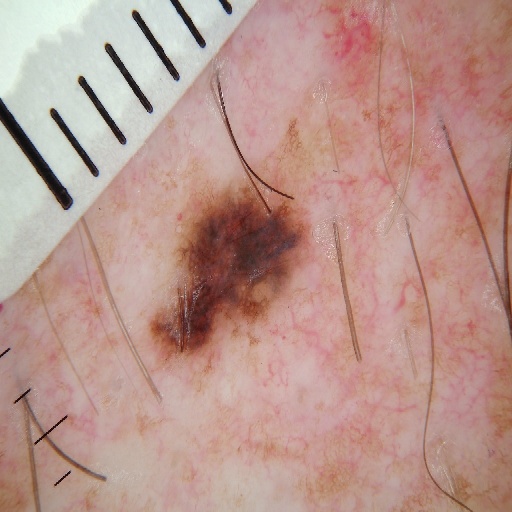}}
		\subfloat{\includegraphics[width=0.84in]{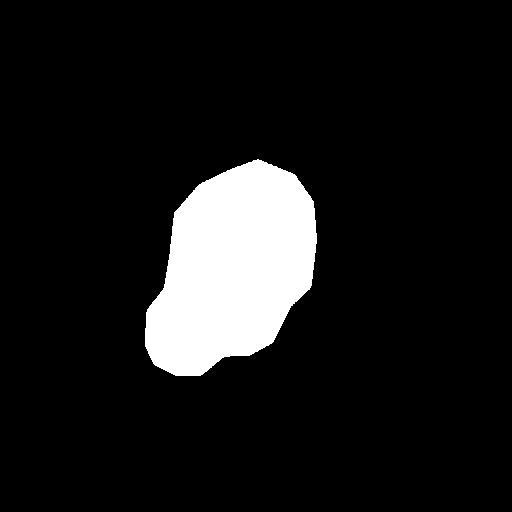}}
		\subfloat{\includegraphics[width=0.84in]{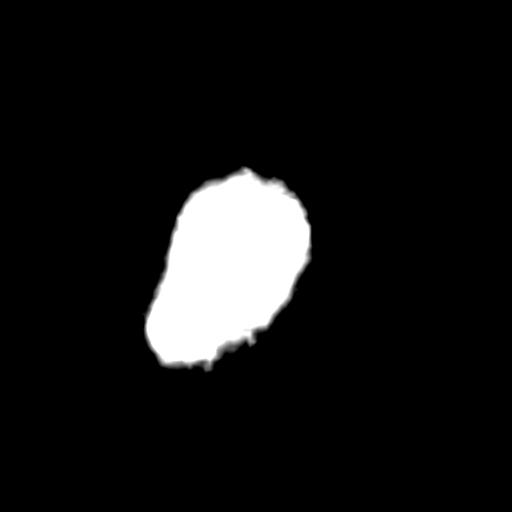}}
		\subfloat{\includegraphics[width=0.84in]{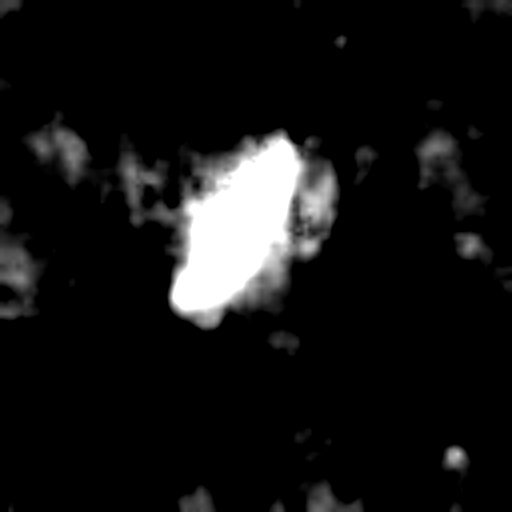}}
		\subfloat{\includegraphics[width=0.84in]{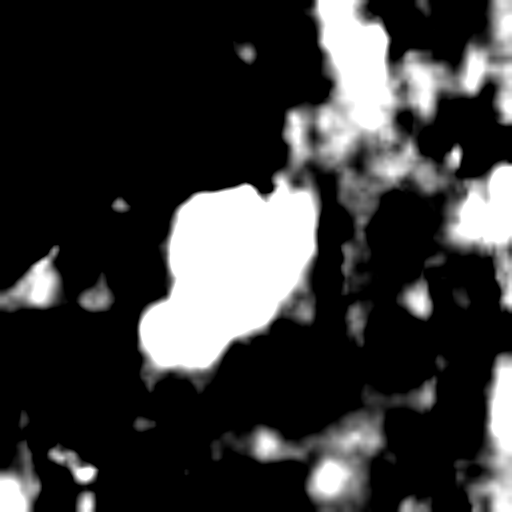}}
		\subfloat{\includegraphics[width=0.84in]{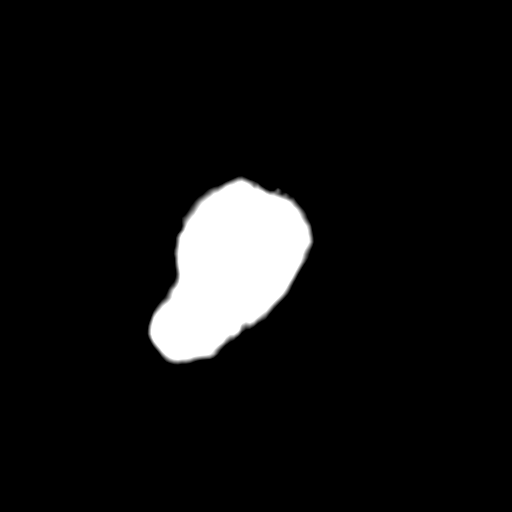}}
		\subfloat{\includegraphics[width=0.84in]{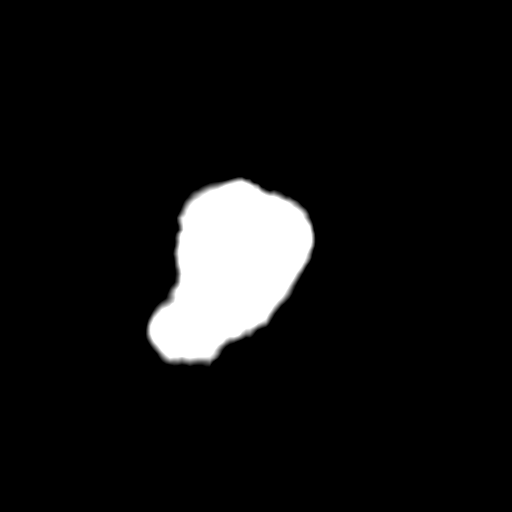}}
		\subfloat{\includegraphics[width=0.84in]{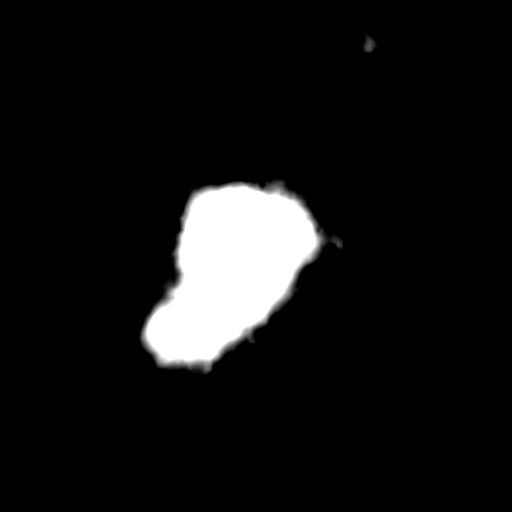}}
		
		\setcounter{subfigure}{0}
		\subfloat[images]{\includegraphics[width=0.84in]{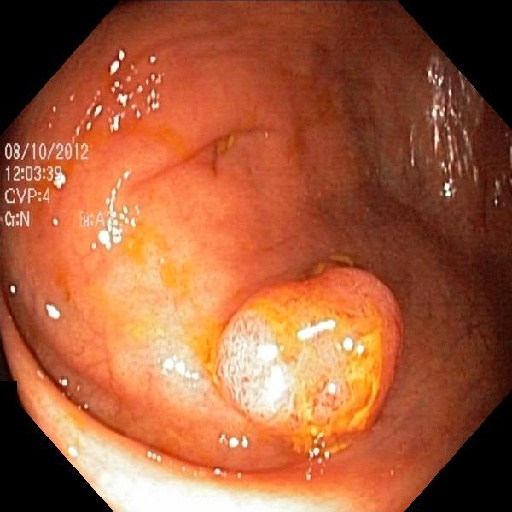}}
		\subfloat[GT]{\includegraphics[width=0.84in]{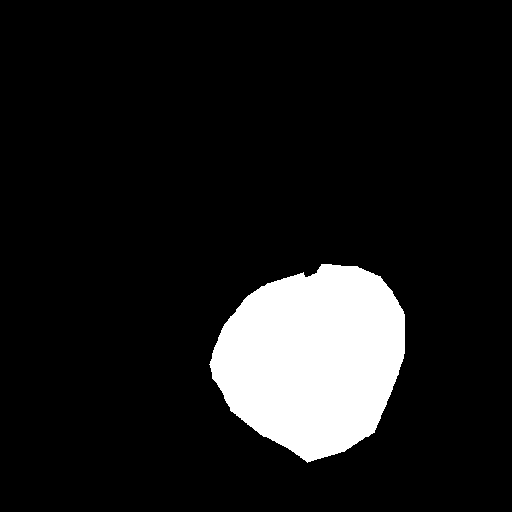}}
		\subfloat[Ours]{\includegraphics[width=0.84in]{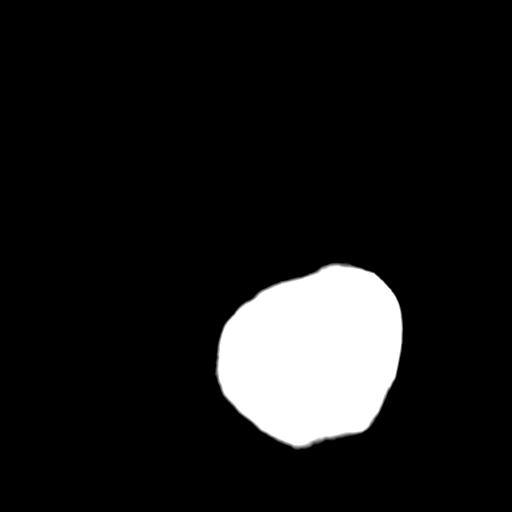}}
		\subfloat[linear probe]{\includegraphics[width=0.84in]{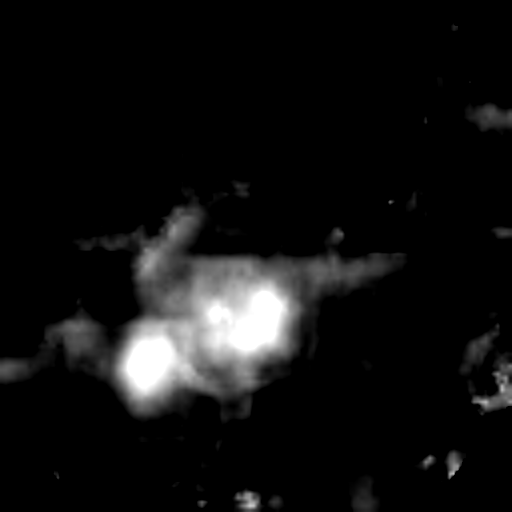}}
		\subfloat[VPT]{\includegraphics[width=0.84in]{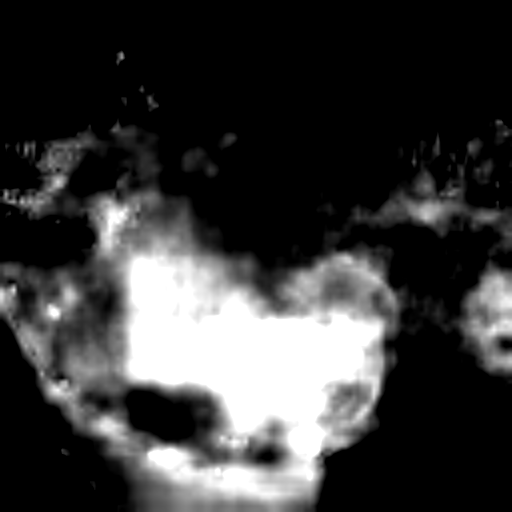}}
		\subfloat[SSF]{\includegraphics[width=0.84in]{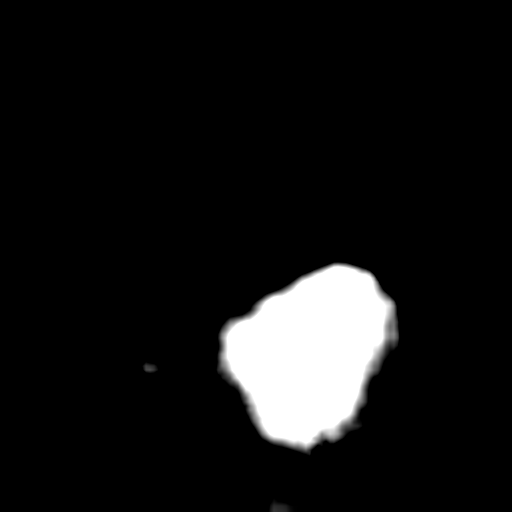}}
		\subfloat[EVP]{\includegraphics[width=0.84in]{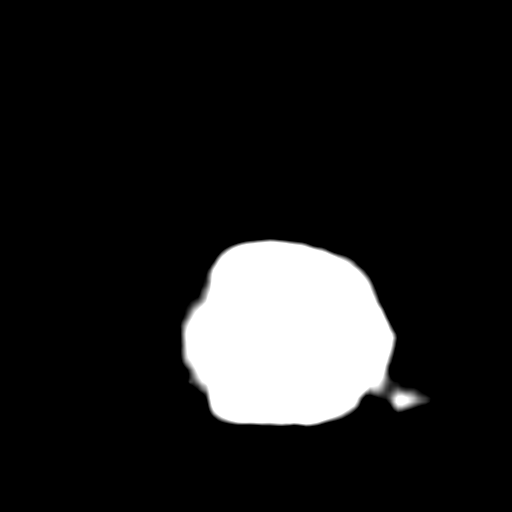}}
		\subfloat[AdaptFormer]{\includegraphics[width=0.84in]{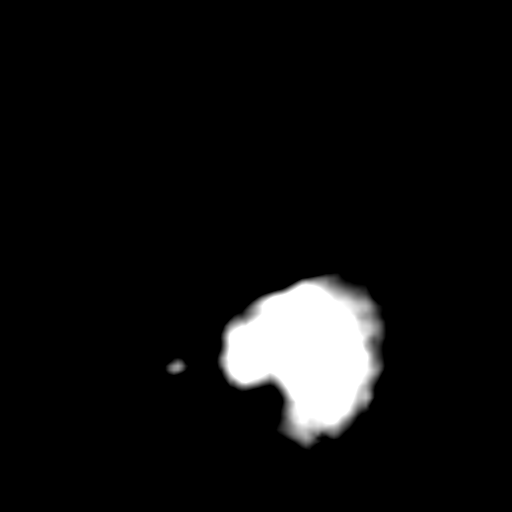}}
		
		\caption{The visual comparison results of different parameter-efficient fine-tuning  methods on skin and polyp segmentation tasks. The images in the first row are the results of skin lesion segmentation, and the second row are the results of polyp segmentation.}
		\label{fig_seg}
	\end{figure*}
	
	\begin{figure}[!t]
		\centering
		\subfloat{\includegraphics[width=3.52in]{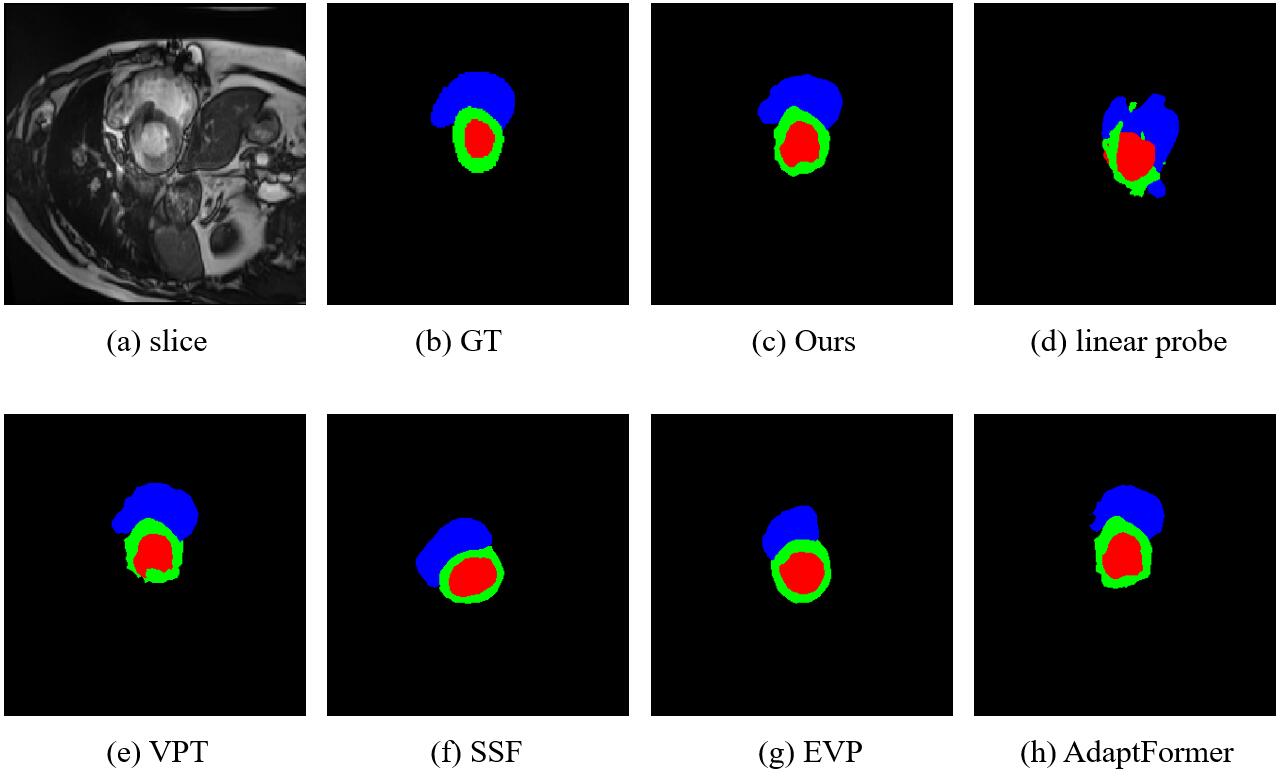}}
		%\subfigure[images]{\includegraphics[width=3.52in]{fig/organ.jpg}}
		\caption{Segmentation results of SOTA PEFT methods for automatic cardiac segmentation on ACDC dataset, red, green, and blue denote the LV, Myo and RV, respectively.} 
		\label{fig_ACDC}
	\end{figure}
	
	\begin{figure}[!t]
		\centering
		\subfloat{\includegraphics[width=3.52in]{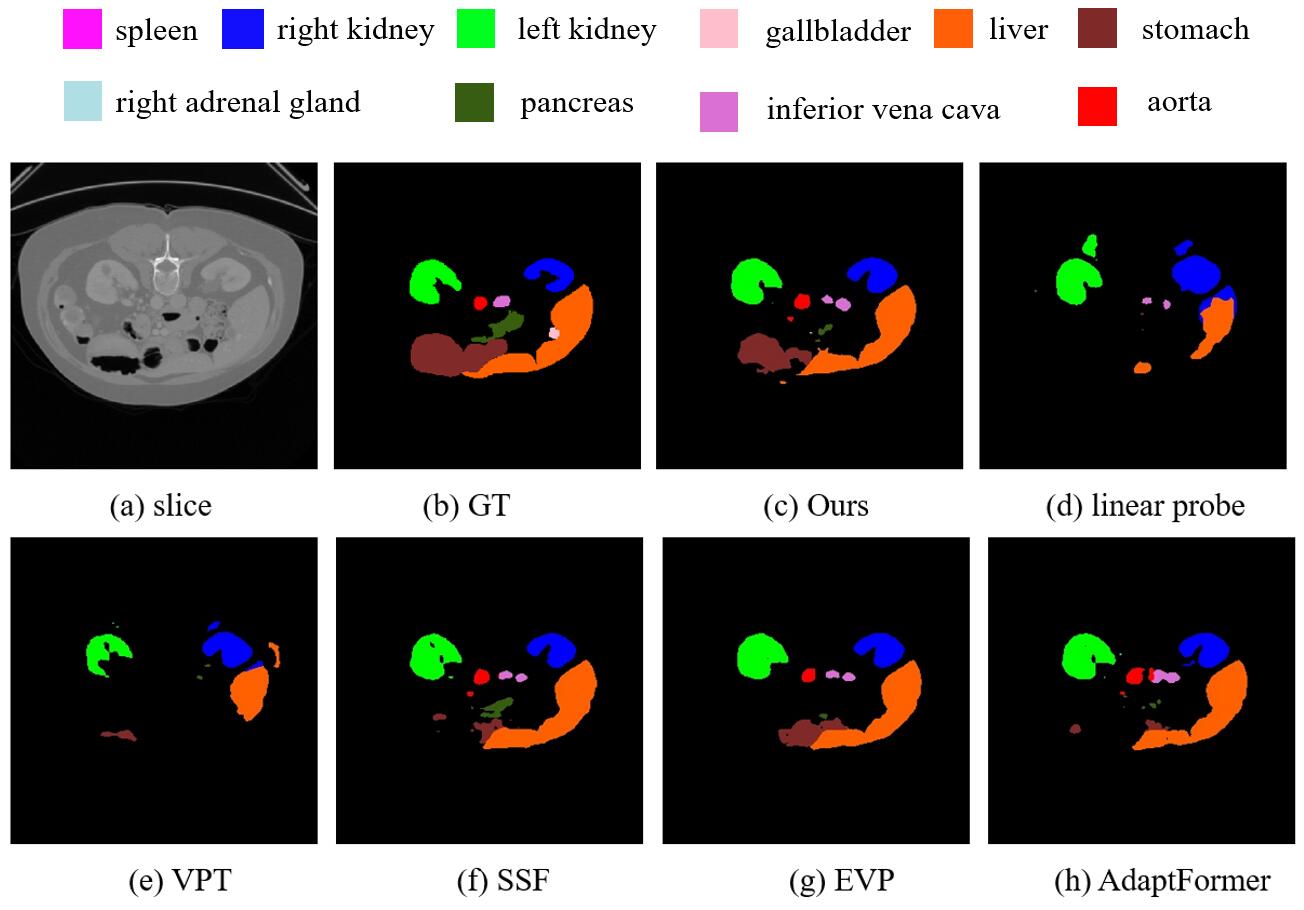}}
		\caption{Segmentation results of SOTA PEFT methods for automatic multi-organ segmentation on Synapse dataset.} 
		\label{fig_organ}
	\end{figure}

	\textbf{The results of segmentation tasks:} To verify the generalization ability of our method, we further conduct experiments on both the 2D and 3D medical image segmentation tasks. The pre-trained large segmentation model is SegFormer-B4 \cite{xie2021segformer} and we report the Dice, IoU and HD95 scores to evaluate the segmentation performance in Table \ref{table_sota_seg}. We can see that linear probe results in very poor performance, and full fine-tuning outperforms linear probe more than 30\% in Dice and IoU scores on polyp, cardiac and multi-organ segmentation, and 8\% on skin lesion segmentation. It can be seen that the performance of VPT is even worse than that of linear probe (theoretical lower bound) on both 2D and 3D segmentation tasks, indicating that VPT is not suitable for segmentation tasks in such a simple visual prompts tuning manner.  Other previous PEFT methods can achieve much better results than linear probe. Then, our method surpasses the previous SOTA PEFT methods on 2D and 3D medical data, which verifies the effectiveness of our method. 
	
	Despite the encouraging results of PEFT methods, the performance of our method is slightly worse than full fine-tuning, it may be due to the fact that the domain shift is more intractable in dense prediction tasks. Finally, we compare our method with commonly used segmentation models and we can see that the large pre-trained model shows strong transfer ability, which can outperform the previous small models by a large margin with very few trainable parameters on 2D medical data. However, it doesn't work as well as the previous SOTA segmentation methods on 3D medical data due to the huge gap between 2D pre-trained data and 3D medical data modalities. We can see that the performance of our method does not degrade much after using the shared DVPT, which verifies adjacent layers have similar feature distributions.
	
	\textbf{Visualization results.} For qualitative analysis of our downstream medical segmentation performance, we show the visual comparisons of our method and other PEFT  methods in Fig. \ref{fig_seg}, Fig. \ref{fig_ACDC} and Fig. \ref{fig_organ}. It can be seen that the predictions of other SOTA methods are susceptible to background noise, and fail to capture the small lesion or organ regions or cannot be able to completely cover the lesion region. From these qualitative results, we find that our method handles the details much better than those of other baselines and has better scalability to regions with different shapes and scales.
	
	\subsection{Ablation Studies}
	\label{ablation}
	We conduct extensive ablation studies to thoroughly evaluate how various designs affect the results, \textit{e.g}., the number of prompts, the hidden dimension of projection layers and the components themselves. The results are shown in Table \ref{table_ablation}.  
	
	\begin{table}
		\centering
		\caption{Ablation studies of our method on DDR dataset with ViT-B/16 backbone. }
		\setlength{\tabcolsep}{3pt}
		\renewcommand\arraystretch{1.2}
		\small
		\subfloat[Ablation studies of DVPT.]{
			\begin{tabular}{l|c|cc}
				\hline
				Settings & \#Params(M)  & Acc(\%) &Kappa(\%)\\
				\hline
				full fine-tuning&          86.2935        & 75.56 &77.68 \\
				\hline
				linear baseline&     0.0025	        & 64.17 &61.62\\
				+ DVPT w/o CAVPT (s=1)&              0.3806         & 73.24 &72.67 \\
				+ DVPT w/ CAVPT (s=1)&               0.4574         &\textbf{76.40} &\textbf{79.91} \\
				\hline
			\end{tabular}
			\label{table_ablation_components}
		}\\
		\subfloat[Ablation study of hidden dimension in projection layers (s=1).]{        
			\begin{tabular}{c|c|c|cc}
				\hline
				hidden dimension  & \#Params(M)  & Acc(\%) &Kappa(\%)\\
				\hline
				5&        0.1365         & 74.67 &76.74 \\
				10&       0.2730         & 75.74 &77.92 \\
				\textbf{20}&       \textbf{0.4574}	     &\textbf{76.40} &\textbf{79.91}\\
				30&       0.6419         & 75.84 &79.03 \\
				40&       0.8263         & 75.74 &78.54 \\
				50&       1.0107         & 74.62 &78.51 \\
				\hline
			\end{tabular}
			\label{table_ablation_dim}
		}\\
		\subfloat[Ablation study of the number of prompts for CAVPT (s=1).]{
			\begin{tabular}{c|c|c|cc}
				\hline
				\#prompts & \#Params(M)  & Acc(\%) &Kappa(\%)\\
				\hline
				30&       0.4267       & 74.83 &76.73 \\
				40&       0.4421	   & 75.95 &77.26\\
				\textbf{50}&       \textbf{0.4574}	     &\textbf{76.40} &\textbf{79.91}\\
				60&       0.4728         & 76.07 &77.94 \\
				70&       0.4881         & 75.46 &76.89 \\
				\hline
			\end{tabular}
			\label{table_ablation_prompt}}
		\label{table_ablation}
	\end{table}
	
	\textbf{Analysis of DVPT.} In Table \ref{table_ablation_components}, we validate the effectiveness of each designs in our method.  We can observe that DVPT without CAVPT can obtains obvious improvements than linear probe, resulting in +9\% and +11\% in Acc and Kappa scores, respectively. When we combine CAVPT with projection layers, the results are already surpass the full fine-tuning method by +2.13\% in Kappa score, further demonstrating the effectiveness of our designs on large pre-trained  vision models with only 0.4574 M trainable parameters. 
	
	\textbf{Analysis of hidden  dimension in projection layers.} The hidden dimension of projection layers controls the number of trainable parameters and we ablate the feature dimension in Table \ref{table_ablation_dim}.  We can see that the performance consistently improves when the middle dimension increases up to 20. As the feature dimension continues to increase, the performance starts to show the degradation. Therefore, we choose the hidden dimension=20 for a better trade-off. The results show that the transformation matrix exhibits low-rank characteristics when performing fine-tuning for large vision models, which is consistent with previous study \cite{hu2021lora}.
	
	\textbf{Analysis of the number of prompts.}  We also investigated the influence of the number of prompts on the final results, as shown in Table \ref{table_ablation_prompt}. Since we only added prompts in the first input layer, the parameters increase slowly as the prompts increase. The best results are achieved when the number of the prompts is 50. We can see that more prompts don't lead to performance improvements, but lead to optimization difficulties. Therefore, we choose 50 as a default setting for the number of trainable prompts.
	
	The above experimental results verify that there is a wealth of knowledge to be mined in the large pre-trained models, and we only need to transfer the knowledge to specific medical tasks with small cost and can achieve competitive results, and this is a new paradigm in the field of medical image analysis. 
\begin{table}[th!] \scriptsize
	\center
	\caption{Analysis of different labeled ratios on downstream datasets. }
	\setlength{\tabcolsep}{3pt}
	\renewcommand\arraystretch{1.2}
	\begin{tabular}{c|cc|cc|cccc}
		\hline
		\multirow{2}{*}{labeled ratio }  & \multicolumn{2}{c|}{ DDR } & \multicolumn{2}{c|}{Polyp} & \multicolumn{2}{c}{Skin}\\
		&Acc(\%)&Kappa(\%)& Dice(\%) & IoU(\%)& Dice(\%) & IoU(\%)\\ 
		\hline
		10\%              & 61.21 &70.31 & 84.81 &77.03& 89.78 &82.73\\
		20\%      	      & 62.35 &74.63 & 88.70 &81.99& 91.03 &84.49\\
		30\%      	      & 64.30 &74.74 & 89.37 &83.22& 91.46 &85.24\\
		40\%              & 74.86 &77.52  & 90.97 &85.35& 91.86 &85.82\\
		\hline
		full fine-tuning  &75.56 &77.68  & \textbf{91.79} &\textbf{86.86}& \textbf{92.88} &\textbf{87.31}\\
		\hline
	\end{tabular}
	\label{table_ablation_ratio}
\end{table}
  \subsection{Efficiency Analysis}
\textbf{Data efficiency:} In order to explore the data efficiency of our parameter-efficient fine-tuning method with different annotation ratios of labeled data, we evaluate our model on downstream datasets. As shown in Table \ref{table_ablation_ratio}, when using only 40\% of the labeled data,  the performance of medical image classification and segmentation tasks can be closer to the results of full fine-tuning with 100\% data. The results demonstrate that our method achieves label-efficient representation learning for medical image analysis, and it can reduce the annotation requirements for downstream tasks by about 60\% without much performance degradation, and thus parameter efficiency brings data efficiency. Our method not only reduces the storage overhead of downstream tasks, but also reduces the dependence on labeled data of downstream tasks, which fully demonstrates the great advantages of our method with large pre-trained models applied to medical downstream tasks, and its strong capability in the low-data regime.

\textbf{Computation and storage efficiency:} To quantitatively analyze the efficiency of the proposed method in terms of computation and storage cost, we report the training memory, FLOPs, and trainable parameters of the SOTA method in Table \ref{table_sota_seg}.  The results are measured in a single NVIDIA GeForce RTX 3090 GPU using the PyTorch framework with the input of 512×512. The fine-tuning time is based on one epoch. Our method saves up to 10.22\% training memory and 99\% storage cost (trainable parameters) of SegFormer-B4 on segmentation task compared to  full fine-tuning. Moreover, DVPT outperforms other SOTA PEFT methods in terms of Dice and IoU with less trainable parameters, and the increased FLOPs of DVPT is negligible. From the results in Table \ref{table_sota_seg}, our model achieves the best trade-off between performance and model training \& storage cost.

\section{Conclusion and Future Work}
\label{Conclusion}
\textbf{Conclusion:} In this work, we focus on  parameter-efficient fine-tuning of pre-trained large vision models for medical image analysis with few trainable parameters, and we propose a novel approach named DVPT. During fine-tuning, we only fine-tune the DVPT and task-specific layers to learn medical knowledge while the pre-trained backbone is fixed. In addition, we can easily control the learning capacity and learnable parameters by altering the number of the  prompts and the shared layers.  DVPT learns sample-specific and domain-specific medical visual representations, reducing the training  computation and storage cost of downstream tasks.  Extensive  experimental results on both 2D and 3D medical classification and segmentation tasks indicate that the proposed  method  outperforms other SOTA methods with a small number of learnable parameters.  It also brings annotation efficiency in comparison to full fine-tuning with all labeled data. It allows the model to be trained efficiently with only 40\% labeled data.  We hope our work inspires  future research in exploring more efficient fine-tuning methods for large pre-trained vision models in medical domain.

\textbf{Limitations and future work:} Due to the limitation of computing resources, our current framework focus on ViT-base and swin-base models, and they are not the largest vision models currently, it can be regarded as a limitation of our work. In the future work, we aim to improve our method to make it suitable for larger model like ViT-large/huge and swin-large/huge models, and we believe that the potential of large pre-trained models can be further unlocked.

{\small
\bibliographystyle{ieee}
\bibliography{egbib}
}

\end{document}